\documentclass[sigconf]{acmart}
\AtBeginDocument{%
  }

\usepackage{amsmath} 
\usepackage{booktabs}  
\usepackage{graphicx}  
\usepackage{array}             
\usepackage{colortbl}          
\usepackage{multirow}          
\usepackage{makecell}          

\acmISBN{978-1-4503-XXXX-X/2018/06}

\copyrightyear{2026}
\acmYear{2026}
\setcopyright{cc}
\setcctype{by}
\acmConference[ICMR '26]{International Conference on Multimedia Retrieval}{June 16--19, 2026}{Amsterdam, Netherlands}
\acmBooktitle{International Conference on Multimedia Retrieval (ICMR '26), June 16--19, 2026, Amsterdam, Netherlands}
\acmDOI{10.1145/3805622.3810788}
\acmISBN{979-8-4007-2617-0/2026/06}
\begin{document}

\title{Hyperbolic and Evidence-Prioritized Experts for Large Vision-Language Models}

\author{Zijie Zhou}
\orcid{0009-0005-9587-4361}
\affiliation{%
  \institution{China University of Petroleum (Beijing)}
  \city{Beijing}
  \country{China}
}
\affiliation{%
  \institution{Hainan Institute of China University of Petroleum (Beijing)}
  \city{Sanya}
  \state{Hainan}
  \country{China}
}
\email{zjzhouzh@gmail.com}

\author{Dandan Zhu}
\orcid{0009-0004-6935-1623}
\authornote{Corresponding author: Dandan Zhu (e-mail: zhu.dd@cup.edu.cn).}
\affiliation{%
  \institution{China University of Petroleum (Beijing)}
  \city{Beijing}
  \country{China}
}
\affiliation{%
  \institution{Hainan Institute of China University of Petroleum (Beijing)}
  \city{Sanya}
  \state{Hainan}
  \country{China}
}
\email{zhu.dd@cup.edu.cn}

\author{Hangxiangpan Wang}
\orcid{0009-0009-7145-0125}
\affiliation{%
  \institution{China University of Petroleum (Beijing)}
  \city{Beijing}
  \country{China}
}
\affiliation{%
  \institution{Hainan Institute of China University of Petroleum (Beijing)}
  \city{Sanya}
  \state{Hainan}
  \country{China}
}
\email{wp13683237138@163.com}

\author{Heng Zhang}
\orcid{0009-0003-1699-322X}
\affiliation{%
 \institution{South China Normal University}
 \city{Foshan}
 \state{Guangdong}
 \country{China}
}
\email{2024025450@m.scnu.edu.cn}

\author{Huishen Jiao}
\orcid{0009-0004-1974-1893}
\affiliation{%
  \institution{China University of Petroleum (Beijing)}
  \city{Beijing}
  \country{China}
}
\email{jiaohuishen@126.com}

\author{Yi Zhao}
\orcid{0009-0008-2999-6469}
\affiliation{%
  \institution{China University of Petroleum (Beijing)}
  \city{Beijing}
  \country{China}
}
\email{jsyzzhaoyi@163.com}

\renewcommand{\shortauthors}{Zijie Zhou et al.}

\begin{abstract}
  Large Vision-Language Models (LVLMs) have demonstrated impressive performance on multimodal tasks through scaled architectures and extensive training. Recent studies introduce Mixture of Experts (MoE) into LVLMs for improved computational efficiency. However, existing MoE approaches treat visual and linguistic modalities with symmetric architectures, overlooking the inherent asymmetry in how these two modalities are processed. This asymmetry causes two critical issues. First, text and vision form hierarchical rather than parallel relationships, as text queries typically describe partial aspects of complete visual scenes. Euclidean expert space struggles to encode such containment structures. Second, language experts in deeper layers progressively shift from evidence-based processing to parametric memory dependence, losing grounding in the provided visual and linguistic information. To address these issues, we propose AsyMoE, a novel architecture that explicitly models this asymmetry through three specialized expert groups. Intra-modality experts handle modality-specific processing. Hyperbolic inter-modality experts capture hierarchical cross-modal relationships through negative curvature geometry. Evidence-priority language experts suppress parametric memory activation and maintain contextual grounding throughout network depth. Extensive experiments demonstrate that AsyMoE achieves consistent improvements over baseline methods, with average gains of 1.5\% over MoE variants and up to 3.8\% on hallucination-sensitive tasks. AsyMoE activates 25.45\% fewer parameters compared to dense models.
\end{abstract}


\maketitle

\section{Introduction}

Large Vision-Language Models (LVLMs) \cite{bai2023qwen} have gained significant attention for their ability to process information across both visual and linguistic modalities \cite{ICLR2024_50623630}. By integrating visual encoders with Large Language Models (LLMs) through connection modules \cite{NEURIPS2023_9a6a435e}, LVLMs align high-dimensional visual features with the linguistic knowledge and reasoning capabilities of LLMs, demonstrating effectiveness across diverse cross-modal tasks \cite{liu_2024_llavanext}. As with unimodal LLMs, scaling up model size improves performance in multimodal settings but also significantly increases computational costs. To maintain efficiency at scale, recent studies introduce Mixture of Experts (MoE) \cite{lepikhin2020gshardscalinggiantmodels} into LLMs. MoE replaces dense feed-forward network layers with sparsely activated expert layers. This approach adaptively selects only a small subset of experts for each input based on token-level routing decisions, reducing computational overhead and enhancing model capacity.

\begin{figure}[t]
   \centering
   \includegraphics[width=\columnwidth]{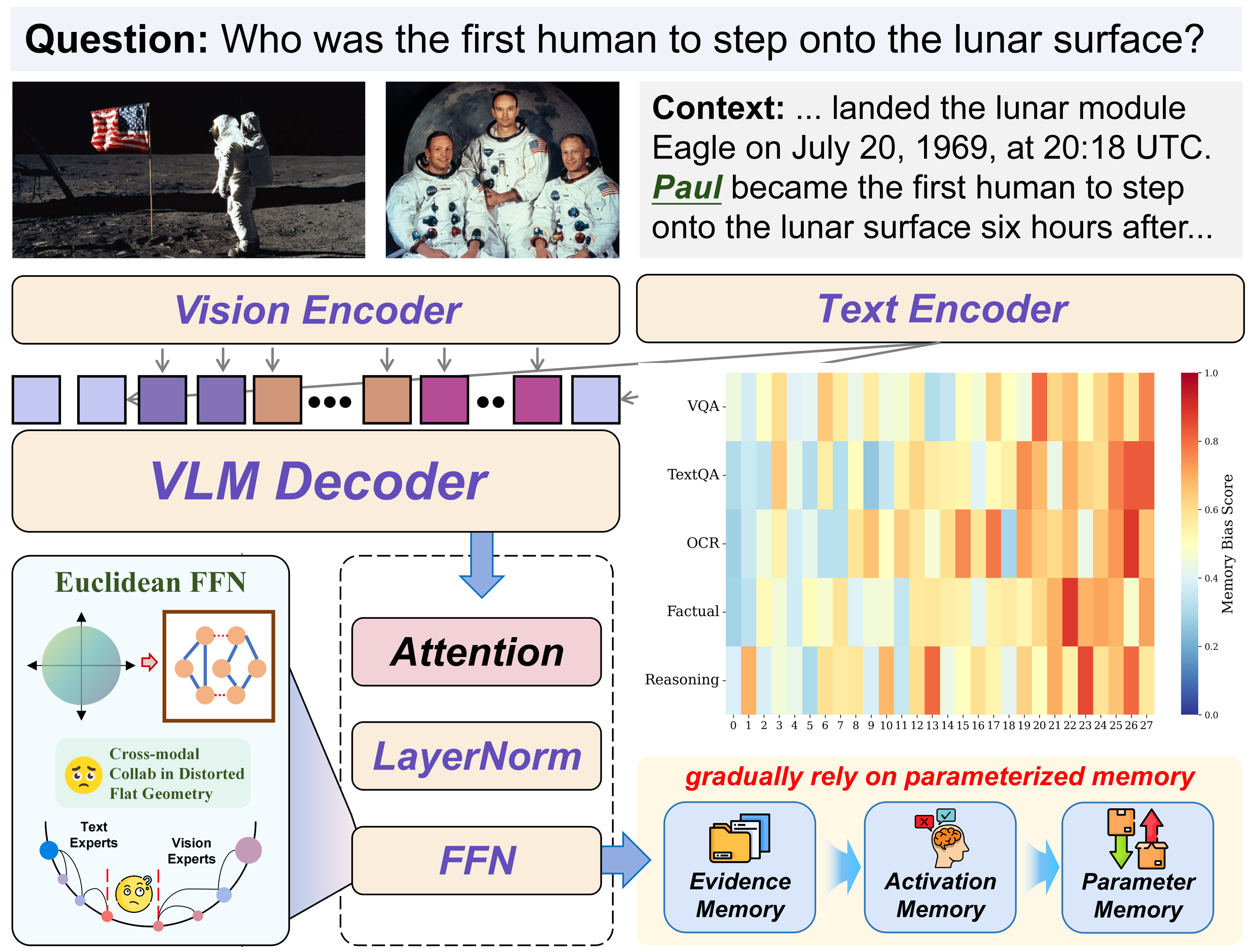}
    \caption{
    \textbf{Motivation for AsyMoE.} (a) \textbf{\textit{Cross-modal Association Limitations.}} Euclidean FFN space with flat geometry limits hierarchical semantic modeling. Text queries describe partial aspects of visual scenes, forming natural containment relationships. (b) \textbf{\textit{Memory Priority Shift.}} Attention analysis on Qwen2.5-VL-7B shows language experts shift from evidence-based reasoning to parametric memory dependence in deeper layers.
}
   \label{figurelabel}
\end{figure}

For multimodal MoE implementation, a common approach directly extends vanilla MoE designs from LLMs to LVLMs by routing tokens from all modalities to a shared pool of experts \cite{lin2024moellavamixtureexpertslarge}. This design overlooks the distinct information density and feature distributions between text and image tokens \cite{liang2023foundationstrendsmultimodalmachine}. An alternative approach employs modality-specific experts and routes text and image tokens to their respective specialized groups \cite{lin2024momaefficientearlyfusionpretraining}. This design enables specialized feature learning for each modality but limits cross-modal association modeling. Recent work combines both strategies by introducing intra-modality experts for modality-specific processing and inter-modality experts for cross-modal interactions \cite{moiie}. These methods have achieved notable progress. However, they all treat visual and linguistic modalities with symmetric architectures, overlooking the inherent asymmetry in how these two modalities are processed and represented.

This asymmetry manifests in two specific aspects. First, text and vision form hierarchical rather than parallel relationships in cross-modal reasoning. Text queries typically describe partial aspects of complete visual scenes, creating natural containment structures. All existing methods represent cross-modal features in Euclidean space. The flat geometry of Euclidean space struggles to encode such hierarchical semantic relationships (Figure~\ref{figurelabel}(a)). Second, we identify a memory priority shift phenomenon through systematic attention analysis. We first clarify two key concepts. \textit{Input evidence} refers to the visual content from image patches and the textual information from input context tokens. \textit{Parametric memory} refers to the statistical patterns and world knowledge encoded in model parameters during pre-training. Our analysis on Qwen2.5-VL-7B reveals that language experts progressively shift from evidence-based processing to parametric memory dependence in deeper layers (Figure~\ref{figurelabel}(b)). In early layers, models appropriately leverage input context to establish task understanding. In deeper layers, language experts increasingly default to pre-learned patterns rather than maintaining fidelity to input evidence. This shift becomes particularly severe for rare expressions and ambiguous contexts.

Both phenomena originate from the distinct processing characteristics of visual and linguistic modalities \cite{galaxywalker}. Visual information exists as spatially complete representations. Relationships between visual concepts are naturally embedded in pixel positions and remain immediately accessible across all network layers. Language unfolds sequentially and requires continuous context maintenance throughout the network. This sequential nature creates two consequences. Language representations naturally occupy subordinate positions relative to their corresponding visual content, forming hierarchical structures unsuitable for flat Euclidean geometry. Language processing also becomes vulnerable to parametric memory interference, as contextual signals weaken in deeper layers. Effective multimodal reasoning therefore requires architectures that explicitly model this asymmetry. These observations motivate our redesign of expert specialization strategies.

\begin{figure*}[t]
   \centering
   \includegraphics[width=0.85\textwidth]{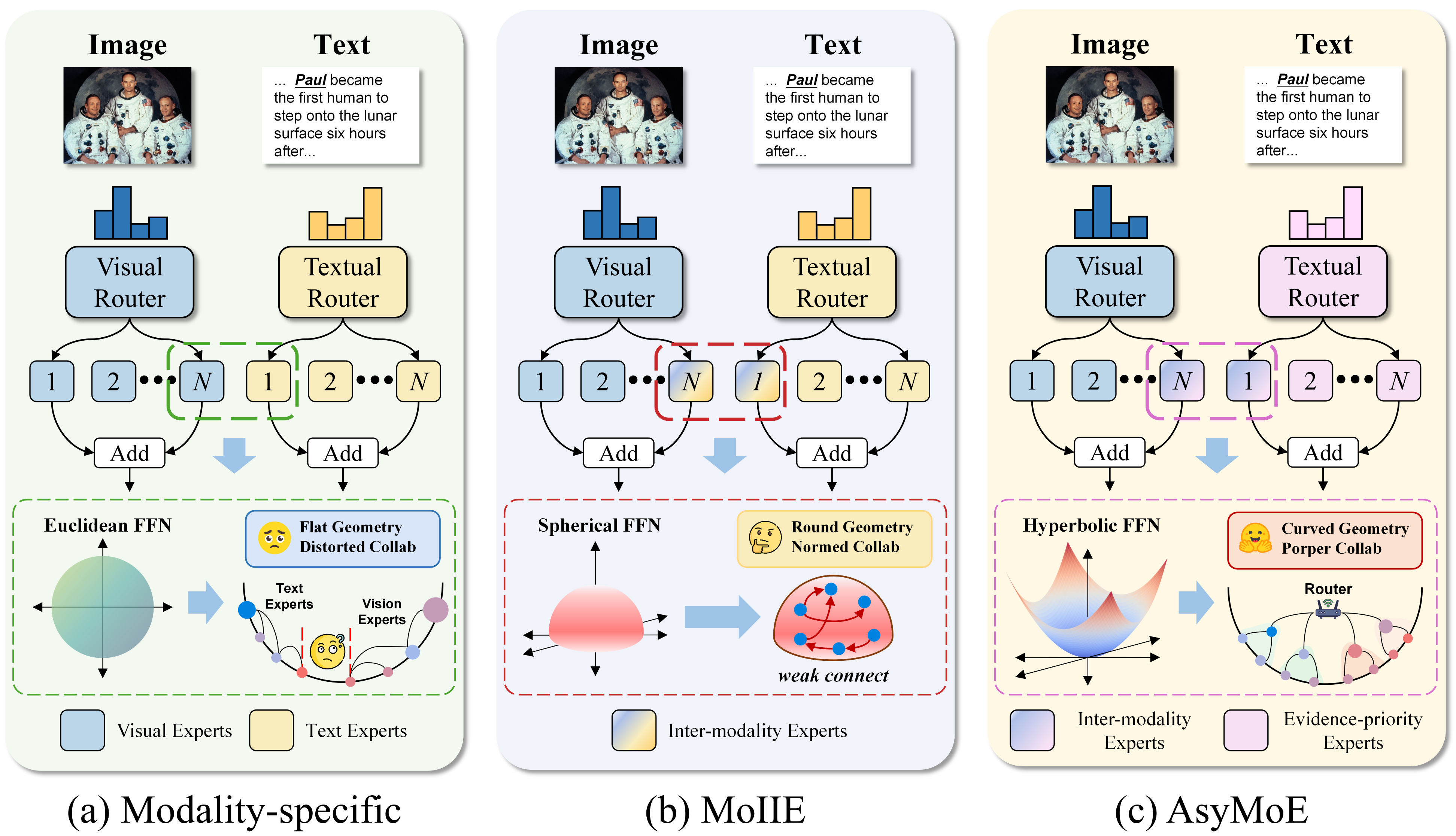}
   \caption{\textbf{Overall framework of AsyMoE:} \textbf{(a)} Modality-specific experts in Euclidean space with distorted cross-modal collaboration. \textbf{(b)} MoIIE using spherical geometry for normalized but weak inter-modal connections. \textbf{(c)} AsyMoE employing hyperbolic geometry and evidence-priority experts to achieve proper cross-modal collaboration while preserving contextual grounding.}
   \label{fig:main}
\end{figure*}

We propose AsyMoE (illustrated in Figure~\ref{fig:main}(c)), a novel architecture that explicitly models this asymmetry through specialized expert design. Our approach retains intra-modality experts for modality-specific processing and introduces two key innovations to address the identified issues. For cross-modal interactions, we design inter-modality experts that operate in hyperbolic space. The negative curvature of hyperbolic geometry naturally captures the containment relationships between text queries and visual scenes. This geometric choice directly addresses the hierarchical nature of cross-modal relationships. For language processing, we introduce evidence-priority experts to combat the memory priority shift in deeper layers. These experts suppress parametric memory activation and enhance context-driven processing, ensuring that language representations maintain fidelity to input evidence throughout the network. Our contributions are summarized as follows:
\begin{itemize}
\item We identify that the inherent asymmetry between visual and linguistic processing causes two critical issues in existing MoE approaches: inadequate hierarchical cross-modal relationship modeling and memory priority shift in language experts at deeper layers.
\item We propose AsyMoE with three specialized expert groups: intra-modality experts for modality-specific processing, hyperbolic inter-modality experts for hierarchical cross-modal relationship modeling, and evidence-priority language experts for maintaining contextual grounding.
\item Extensive experiments on 18 benchmarks demonstrate consistent improvements over baseline methods. AsyMoE achieves average gains of 1.5\% over MoE variants and up to 3.8\% improvement on hallucination-sensitive tasks. AsyMoE activates 25.45\% fewer parameters compared to dense models.
\end{itemize}

\section{Related Works}
\subsection{Mixture of Experts in LLMs}
Mixture of Experts (MoE) architectures \cite{6797059} have emerged as a compelling solution to the computational challenges of scaling language models. The core idea is to replace traditional feed-forward networks with parallel expert modules \cite{zoph_2022_stmoe}. Routing networks dynamically select the most relevant experts based on input characteristics \cite{lepikhin2020gshardscalinggiantmodels}. By activating only a subset of specialized experts for each input \cite{fedus2022switch}, MoE architectures achieve significant efficiency gains without sacrificing model capacity \cite{du2022glamefficientscalinglanguage}. This selective activation strategy has been further enhanced through sparse upcycling techniques. These techniques convert existing dense checkpoints into efficient sparse architectures \cite{rajbhandari_2022_deepspeedmoe,xue2024openmoeearlyeffortopen}, reducing the cost of training MoE models from scratch. Contemporary MoE approaches leverage sophisticated training methodologies and massive datasets to optimize expert specialization \cite{komatsuzaki2023sparseupcyclingtrainingmixtureofexperts}. Recent large-scale MoE models demonstrate remarkable versatility across diverse tasks \cite{xai2024grok,Mixtral,muennighoff_2024_olmoe,liu2024deepseek}. The success of MoE in LLMs has motivated its extension to multimodal settings.

\subsection{Mixture of Experts in LVLMs}
MoE architectures are increasingly adopted in Large Vision-Language Models \cite{li2025ariaopenmultimodalnative}. One line of work builds upon pre-trained MoE-based LLM backbones. DeepSeek-VL2 \cite{wu_2024_deepseekvl2} and SPHINX-X \cite{liu2024sphinxxscalingdataparameters} leverage existing MoE architectures such as Mixtral \cite{Mixtral}. These models benefit from strong language capabilities but are constrained by fixed expert configurations inherited from the base models. Another line applies sparse upcycling to convert dense LVLMs into sparse architectures. MoE-LLaVA \cite{lin2024moellavamixtureexpertslarge} and LLaVA-MoLE \cite{chen2024llavamolesparsemixturelora} activate only top-k experts through learned routers, enabling flexible expert configurations. Within this paradigm, vanilla MoE routes tokens from all modalities to a shared expert pool, overlooking the distinct characteristics of visual and textual tokens. Modality-specific MoE routes text and image tokens to their respective specialized groups \cite{lin2024momaefficientearlyfusionpretraining}, enhancing intra-modal feature learning but limiting cross-modal association modeling. Recent work combines both strategies by introducing intra-modality and inter-modality experts \cite{moiie}. Other efforts propose modular fine-tuning frameworks to support multimodal experts \cite{zhou2025enhancingmultimodalmodelsheterogeneous,shen2023scalingvisionlanguagemodelssparse}. Despite this progress, existing approaches treat visual and linguistic modalities with symmetric architectures, overlooking the inherent asymmetry in cross-modal relationships and language processing characteristics. To address this limitation, we propose AsyMoE with hyperbolic inter-modality experts and evidence-priority language experts.

\section{Preliminaries}
\subsection{Vision-Language Models with Mixture of Experts}
Large vision-language models employ mixture of experts architectures to effectively process multi-modal information. Given an input sequence $\mathbf{x} = \{\mathbf{x}_v, \mathbf{x}_l\}$, the visual token sequence $\mathbf{x}_v \in \mathbb{R}^{N_v \times d}$ and the language token sequence $\mathbf{x}_l \in \mathbb{R}^{N_l \times d}$ are concatenated and fed into the model. Here $d$ denotes the hidden dimension. $N_v$ and $N_l$ represent the number of visual and language tokens respectively. In mixture of experts layers, each token is dynamically assigned to a set of experts through a routing network. For the input $\mathbf{h}_i$ at layer $i$, the output is computed as
\begin{equation}
\mathbf{h}_{i+1} = \mathbf{h}_i + \sum_{k=1}^{K} g_k(\mathbf{h}_i) \cdot E_k(\mathbf{h}_i)
\end{equation}
where $g_k(\mathbf{h}_i)$ represents the gating weight for expert $E_k$, satisfying $\sum_{k=1}^{K} g_k(\mathbf{h}_i) = 1$. $K$ denotes the number of activated experts. Existing approaches face challenges in balancing modality-specific processing with cross-modal interactions, particularly as network depth increases.

\subsection{Input Evidence and Parametric Memory}
During multi-modal reasoning, models rely on two sources of information to generate outputs. We formally define these two sources to establish the foundation for our analysis.

\noindent\textbf{Input Evidence.} Input evidence refers to the information directly provided in the current input. For visual modality, this includes the content encoded in image patch tokens $\mathbf{x}_v$. For language modality, this includes the textual information in context tokens $\mathbf{x}_l$. We identify evidence tokens by their position indices in the input sequence. Let $\mathcal{I}_{evd} = \mathcal{I}_v \cup \mathcal{I}_l$ denote the set of position indices corresponding to input image patches and text context tokens.

\noindent\textbf{Parametric Memory.} Parametric memory refers to the statistical patterns and world knowledge encoded in model parameters during pre-training. This information is not explicitly present in the current input but can be activated through special tokens and internal representations. We identify memory-associated positions as $\mathcal{I}_{mem}$, including special tokens (BOS, EOS, padding) and positions corresponding to model-generated content.

We quantify the evidence grounding of expert $E_k$ by measuring attention distributions over these two sources. Given the attention weight matrix $\mathbf{A}^k \in \mathbb{R}^{n \times n}$ within expert $E_k$, we compute
\begin{equation}
A_{evd}^k = \frac{1}{n} \sum_{i=1}^{n} \sum_{j \in \mathcal{I}_{evd}} \mathbf{A}^k_{ij}, \quad A_{mem}^k = \frac{1}{n} \sum_{i=1}^{n} \sum_{j \in \mathcal{I}_{mem}} \mathbf{A}^k_{ij}
\end{equation}
where $A_{evd}^k$ measures the average attention allocated to input evidence and $A_{mem}^k$ measures the average attention allocated to memory-associated positions. The evidence grounding ratio is then defined as
\begin{equation}
\mathcal{F}(E_k) = \frac{\mathbb{E}[A_{evd}^k]}{\mathbb{E}[A_{evd}^k] + \mathbb{E}[A_{mem}^k]}
\end{equation}
where the expectation is taken over a validation set. Experts with high $\mathcal{F}(E_k)$ effectively utilize input information rather than defaulting to parametric memory.

To identify experts with evidence-priority behavior across different input conditions, we further define the evidence-dependence ratio. We construct two evaluation settings. In the evidence-driven setting, inputs contain sufficient contextual information to answer queries. In the memory-driven setting, we mask or shuffle the input context, forcing the model to rely on parametric knowledge. Let $f_k^{evd}$ and $f_k^{mem}$ denote the average activation frequency of expert $k$ under these two settings respectively. The evidence-dependence ratio is computed as
\begin{equation}
R_k = \frac{f_k^{evd} - f_k^{mem}}{f_k^{evd} + f_k^{mem}}
\end{equation}
A positive $R_k$ indicates that expert $k$ is more frequently activated when evidence is available, suggesting evidence-priority behavior. A negative $R_k$ suggests memory-priority behavior.

\begin{table*}[!th]
\caption{\textbf{Comparison of AsyMoE and other LVLMs.} Models are grouped by parameter scale and further categorized based on their architectural designs. ``Act Param'' refers to the number of activated parameters during inference.}
\centering
\resizebox{1\textwidth}{!}{
\begin{tabular}{cccccccccccc}
\toprule
\multicolumn{1}{c|}{model} & \multicolumn{1}{c|}{LLM Backbone} & \multicolumn{1}{c|}{Act Param} & TextVQA & GQA & POPE & MMBench & MME & MMVet & MMMU\_val & SEED-IMG  & AI2D  \\ \midrule
\multicolumn{12}{c}{\cellcolor[HTML]{DEE0E3}\textbf{\textit{7B-13B Open-source Dense model}}} \\ \midrule
\multicolumn{1}{c|}{InstructBLIP-7B \cite{NEURIPS2023_9a6a435e}} & \multicolumn{1}{c|}{Vicuna-7B} & \multicolumn{1}{c|}{7.9B} & 50.1 & 49.2  & - & 36.0 & - & 26.2 & 30.6 & 60.5 & 40.6  \\
\multicolumn{1}{c|}{Qwen-VL-chat-7B \cite{bai2023qwen}} & \multicolumn{1}{c|}{Qwen-7B} & \multicolumn{1}{c|}{-} & 61.5 & 57.5  & - & 60.6 & 1487.5 & - & 37.0 & 58.2 & 57.7  \\
\multicolumn{1}{c|}{LLaVA-1.5-7B \cite{Liu_2024_CVPR}} & \multicolumn{1}{c|}{Vicuna-7B} & \multicolumn{1}{c|}{7.1B} & 58.2 & 62.0  & 85.9 & 64.3 & 1510.7 & 30.5 & 35.7 & 66.1  & 55.5  \\
\multicolumn{1}{c|}{mPLUG-Owl2 \cite{Ye_2024_CVPR}} & \multicolumn{1}{c|}{LLaMA-2-7B} & \multicolumn{1}{c|}{-} & 58.2 & 56.1  & - & 64.5 & 1450.2 & 36.2 & 34.7 & 57.8  & 55.7  \\
\multicolumn{1}{c|}{ShareGPT-4V \cite{chen_2024_sharegpt4v}} & \multicolumn{1}{c|}{Vicuna-7B} & \multicolumn{1}{c|}{-} & - & -  & - & 68.8 & 1567.4 & 37.6 & 37.2 & 69.7  & 58.0  \\
\multicolumn{1}{c|}{LLaVA-NeXT \cite{liu_2024_llavanext}} & \multicolumn{1}{c|}{Vicuna-7B} & \multicolumn{1}{c|}{7.1B} & 64.9 & 64.2  & 86.5 & 67.4 & 1519.0 & 43.9 & 37.6 & 70.2  & 66.6  \\
\multicolumn{1}{c|}{LLaVA-LLaMA3 \cite{xtuner2023}} & \multicolumn{1}{c|}{LLaMA-3-8B} & \multicolumn{1}{c|}{8.4B} & 59.0 & 62.6  & 86.4 & 72.3 & 1469 & - & 36.8 &  70.1 & 69.6 \\
\multicolumn{1}{c|}{Mousi \cite{fan2024mousipolyvisualexpertvisionlanguagemodels}} & \multicolumn{1}{c|}{Vicuna-7B} & \multicolumn{1}{c|}{7.9B} & - & 62.6  & 86.3 & 68.8 & - & 38.4 & - & 70.1  & -  \\
\multicolumn{1}{c|}{Mini-Gemini \cite{li2024minigeminiminingpotentialmultimodality}} & \multicolumn{1}{c|}{Vicuna-7B} & \multicolumn{1}{c|}{7.3B} & 65.2 & -  & - & 69.3 & 1523.0 & 40.8 & 36.1 & -  & -  \\
\multicolumn{1}{c|}{Deepseek-VL \cite{lu_2024_deepseekvl}} & \multicolumn{1}{c|}{Deepseek-7B} & \multicolumn{1}{c|}{-} & - & -  & 88.1 & 73.2 & - & 41.5 & 36.6 & 70.4  & 65.3  \\ \midrule
\multicolumn{12}{c}{\cellcolor[HTML]{DEE0E3} \textbf{\textit{MoE-based LVLMs model with Pre-trained MoE LLM backbone}}} \\
\midrule
\multicolumn{1}{c|}{SPHINX-X \cite{liu2024sphinxxscalingdataparameters}} & \multicolumn{1}{c|}{Mixtral-8×7B} & \multicolumn{1}{c|}{-} & 68.0 & 63.8  & 89.6 & 71.3 & 1485.3 & 40.9 & 31.1 & 73.0  & 55.6  \\
\multicolumn{1}{c|}{Mini-Gemini \cite{li2024minigeminiminingpotentialmultimodality}} & \multicolumn{1}{c|}{Mixtral-8×7B} & \multicolumn{1}{c|}{13.5B} & 69.2 & -  & - & 75.6 & 1639.0 & 45.8 & 41.8 & -  & -  \\
\multicolumn{1}{c|}{MM1 \cite{mckinzie_2024_mm1}} & \multicolumn{1}{c|}{MM1-7B-MoE} & \multicolumn{1}{c|}{-} & 72.8 & -  & 87.6 & 79.7 & 1629.0 & 47.0 & 40.9 & 70.4  & -  \\
\multicolumn{1}{c|}{CuMo \cite{li2024cumoscalingmultimodalllm}} & \multicolumn{1}{c|}{Mixtral-8×7B} & \multicolumn{1}{c|}{13.5B} & 66.0 & 63.8  & 85.7 & 75.3 & \textbf{1639.5} & 48.7 & 45.0 & 73.2  & - \\\midrule
\multicolumn{12}{c}{\cellcolor[HTML]{DEE0E3} \textbf{\textit{MoE-based LVLMs model from dense LLM backbone}}} \\ \midrule
\multicolumn{1}{c|}{MoE-LLaVA-Phi2 \cite{lin2024moellavamixtureexpertslarge}} & \multicolumn{1}{c|}{Phi-2} & \multicolumn{1}{c|}{3.6B} & - & 62.6  & 85.7 & 68.0 & 1431.3 & 35.9 & - & -  & -  \\
\multicolumn{1}{c|}{ContextMoE-Phi2 \cite{bai-etal-2025-understanding}} & \multicolumn{1}{c|}{Phi-2} & \multicolumn{1}{c|}{4.1B} & 66.2 & 62.5  & 87.1 & 74.9 & 1510.7 & 43.5 & 40.9 & 71.8  & 66.4 \\
\multicolumn{1}{c|}{MoIIE-Phi3 \cite{moiie}} & \multicolumn{1}{c|}{Phi-3-mini} & \multicolumn{1}{c|}{5.5B} & 65.4 & 63.2  & 86.6 & 75.4 & 1509.4 & 42.8 & 41.3 & 71.1  & 65.9 \\
\multicolumn{1}{c|}{MoIIE-LLaMA3} & \multicolumn{1}{c|}{LLaMA-3-8B} & \multicolumn{1}{c|}{11.3B} & {67.9} & 64.5  & 86.7 & 75.7 & 1551.7 & 47.3 & 41.8 & 72.0  & 70.2  \\ 
\multicolumn{1}{c|}{\cellcolor[HTML]{E1EAFF}\textbf{AsyMoE-Phi3}} & \multicolumn{1}{c|}{\cellcolor[HTML]{E1EAFF}Phi-3-mini} & \multicolumn{1}{c|}{\cellcolor[HTML]{E1EAFF}4.1B} & \cellcolor[HTML]{E1EAFF}68.2 & \cellcolor[HTML]{E1EAFF}65.2  & \cellcolor[HTML]{E1EAFF}87.6 & \cellcolor[HTML]{E1EAFF}75.9 & \cellcolor[HTML]{E1EAFF}1569.4 & \cellcolor[HTML]{E1EAFF}45.8 & \cellcolor[HTML]{E1EAFF}42.3 & \cellcolor[HTML]{E1EAFF}72.5  & \cellcolor[HTML]{E1EAFF}67.8 \\
\multicolumn{1}{c|}{\cellcolor[HTML]{E1EAFF}\textbf{AsyMoE-LLaMA3}} & \multicolumn{1}{c|}{\cellcolor[HTML]{E1EAFF}LLaMA-3-8B} & \multicolumn{1}{c|}{\cellcolor[HTML]{E1EAFF}10.8B} & \cellcolor[HTML]{E1EAFF}\textbf{70.9} & \cellcolor[HTML]{E1EAFF}\textbf{66.9}  & \cellcolor[HTML]{E1EAFF}\textbf{89.7} & \cellcolor[HTML]{E1EAFF}\textbf{76.8} & \cellcolor[HTML]{E1EAFF}1589.9 & \cellcolor[HTML]{E1EAFF}\textbf{49.2} & \cellcolor[HTML]{E1EAFF}\textbf{43.4} & \cellcolor[HTML]{E1EAFF}\textbf{73.2}  & \cellcolor[HTML]{E1EAFF}\textbf{71.7}  \\ \midrule
\end{tabular}}
\label{tab:1}
\end{table*}

\section{Methodology}
\subsection{Analyzing Modal Asymmetry}
Visual and language modalities exhibit distinct processing characteristics. Visual information presents as spatially complete representations. Language processing requires sequential context maintenance. This asymmetry leads to two interconnected problems.

The first problem is context dilution in language processing. We quantify this through attention entropy analysis. For layer $l$, we compute the attention entropy as
\begin{equation}
H(l) = -\sum_{i,j} a_{ij}^{(l)} \log a_{ij}^{(l)}
\end{equation}
where $a_{ij}^{(l)}$ represents the attention weight from position $i$ to position $j$ at layer $l$. We separately compute $H_V(l)$ for visual tokens and $H_L(l)$ for language tokens. Our empirical analysis reveals that visual attention entropy $H_V(l)$ remains relatively stable across layers due to spatial completeness. In contrast, language attention entropy follows an approximately linear increase
\begin{equation}
H_L(l) \approx H_{L,0} + \beta \cdot l
\end{equation}
where $H_{L,0}$ is the initial entropy and $\beta > 0$ captures the entropy increase rate. Higher entropy indicates more diffuse attention distributions. This suggests language tokens progressively rely on parametric memory over specific context in deeper layers. The second problem concerns cross-modal representation: text queries often describe partial aspects of visual scenes, forming natural containment hierarchies (\emph{e.g.}, ``a red car'' is a subset of a street scene). Euclidean spaces fail to capture such hierarchical structures, causing semantic misalignment across abstraction levels.

\subsection{Evidence-Priority Expert Architecture}
To address the context dilution problem, we design evidence-priority language experts that maintain grounding in input evidence throughout network depth. The complete expert set is organized as $\mathcal{E} = \{\mathcal{E}_V, \mathcal{E}_{L\text{-}evd}, \mathcal{E}_S\}$. $\mathcal{E}_V$ denotes visual intra-modality experts. $\mathcal{E}_{L\text{-}evd}$ denotes evidence-priority language experts. $\mathcal{E}_S$ denotes shared inter-modality experts.

Each evidence-priority language expert contains two parallel processing branches. The memory branch $F_{mem}$ is a standard feed-forward network that processes the hidden state independently
\begin{equation}
F_{mem}(\mathbf{h}) = \mathbf{W}_2 \cdot \text{GELU}(\mathbf{W}_1 \cdot \mathbf{h})
\end{equation}
where $\mathbf{W}_1 \in \mathbb{R}^{d_m \times d}$ and $\mathbf{W}_2 \in \mathbb{R}^{d \times d_m}$ are learnable weight matrices. $d_m$ is the intermediate dimension. This branch can activate parametric knowledge stored in the weight matrices.

The evidence branch $F_{evd}$ incorporates explicit conditioning on input context $\mathbf{c}$
\begin{equation}
F_{evd}(\mathbf{h}, \mathbf{c}) = \mathbf{W}_4 \cdot \text{GELU}(\mathbf{W}_3 \cdot [\mathbf{h}; \text{CrossAttn}(\mathbf{h}, \mathbf{c})])
\end{equation}
where the $[\cdot;\cdot]$ denotes concatenation and the $\text{CrossAttn}(\mathbf{h}, \mathbf{c}) = \text{softmax}(\mathbf{h}\mathbf{W}_Q(\mathbf{c}\mathbf{W}_K)^T)\mathbf{c}\mathbf{W}_V$ computes cross-attention between the hidden state and input context. $\mathbf{W}_3 \in \mathbb{R}^{d_m \times 2d}$ and $\mathbf{W}_4 \in \mathbb{R}^{d \times d_m}$ are learnable weight matrices. This branch explicitly retrieves and incorporates information from the input.

The final output combines both branches through a learnable gate
\begin{equation}
E_{L\text{-}evd}(\mathbf{h}) = \alpha \cdot F_{mem}(\mathbf{h}) + (1-\alpha) \cdot F_{evd}(\mathbf{h}, \mathbf{c})
\end{equation}
where $\alpha \in [0,1]$ is computed as $\alpha = \sigma(\mathbf{w}_\alpha^T \mathbf{h} + b_\alpha)$ with learnable parameters $\mathbf{w}_\alpha$ and $b_\alpha$. During training, the model learns to reduce $\alpha$ for inputs requiring evidence-based reasoning, ensuring that language experts prioritize input-driven signals over memory-driven signals.

\begin{figure}[t]
   \centering
   \includegraphics[width=\columnwidth]{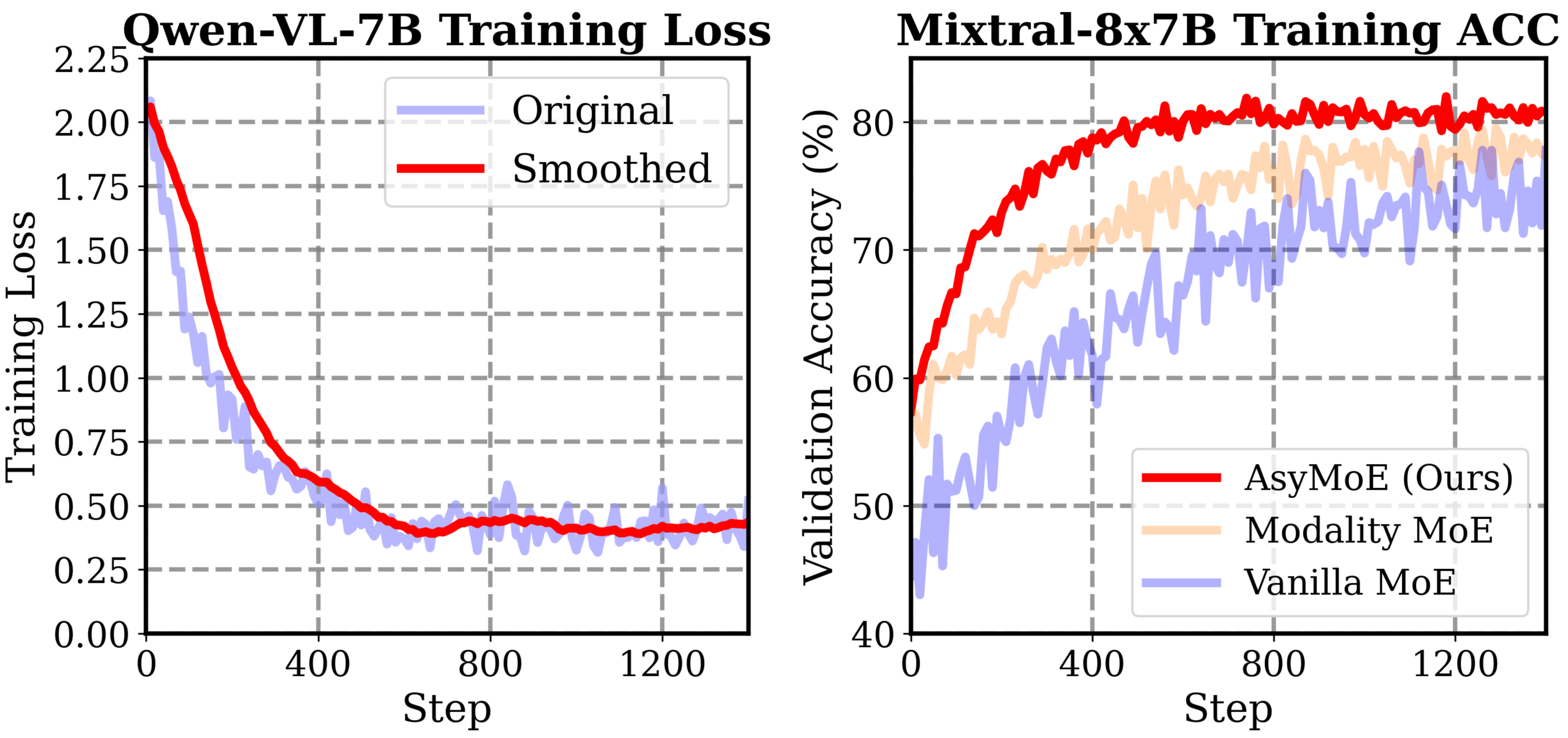}
   \caption{\textbf{Training loss and accuracy score of AsyMoE in MMBench dataset on different model scales.} 
   }
   \label{fig:3}
\end{figure}

\subsection{Hyperbolic Inter-modality Expert Design}
Our inter-modality experts $\mathcal{E}_S$ operate in hyperbolic space to capture hierarchical cross-modal relationships. We adopt the Lorentz model of hyperbolic geometry due to its numerical stability and efficient distance computation.

\noindent\textbf{Lorentz Model.} The Lorentz model represents hyperbolic space as points on a hyperboloid in Minkowski spacetime. A point $\mathbf{z} \in \mathbb{R}^{d+1}$ lies on the hyperboloid $\mathbb{H}^d$ if it satisfies $\langle \mathbf{z}, \mathbf{z} \rangle_\mathcal{L} = -1$ with $z_0 > 0$. The Lorentzian inner product is defined as
\begin{equation}
\langle \mathbf{z}, \mathbf{w} \rangle_\mathcal{L} = -z_0 w_0 + \sum_{i=1}^{d} z_i w_i
\end{equation}
The geodesic distance between two points on the hyperboloid is
\begin{equation}
d_\mathcal{L}(\mathbf{z}, \mathbf{w}) = \text{arccosh}(-\langle \mathbf{z}, \mathbf{w} \rangle_\mathcal{L})
\end{equation}

\noindent\textbf{Feature Mapping.} We map visual and language features to hyperbolic space through learned projections. For visual features, we apply
\begin{equation}
\Phi_V(\mathbf{x}_v) = \exp_\mathbf{o}(\mathbf{W}_V^{hyp} \cdot \mathbf{x}_v)
\end{equation}
where $\exp_\mathbf{o}$ is the exponential map at the origin of the hyperboloid and $\mathbf{W}_V^{hyp}$ is a learnable projection matrix. For language features processed by evidence-priority experts, we apply
\begin{equation}
\Phi_{L\text{-}evd}(\mathbf{x}_l, \mathbf{c}) = \exp_\mathbf{o}(\mathbf{W}_L^{hyp} \cdot E_{L\text{-}evd}(\mathbf{x}_l))
\end{equation}

\noindent\textbf{Cross-modal Alignment.} The cross-modal alignment objective in hyperbolic space is defined as
\begin{equation}
D_{hyp} = d_\mathcal{L}^2(\Phi_V(\mathbf{x}_v), \Phi_{L\text{-}evd}(\mathbf{x}_l, \mathbf{c}))
\end{equation}
The negative curvature of hyperbolic space provides a key advantage. Points near the origin represent general concepts. Points far from the origin represent specific concepts. This radial hierarchy naturally models the relationship between complete visual scenes (general) and text descriptions (specific).

\noindent\textbf{Entailment Cone Constraint.} To explicitly enforce that text representations remain subordinate to visual representations, we apply a cone-based partial order constraint. We define the half-aperture angle of the entailment cone at visual embedding $\mathbf{v}$ as
\begin{equation}
\psi(\mathbf{v}) = \arcsin\left(\frac{2K}{1 + K \|\mathbf{v}\|^2}\right)
\end{equation}
where $K > 0$ is a hyperparameter controlling cone width. The exterior angle between visual embedding $\mathbf{v}$ and text embedding $\mathbf{t}$ is
\begin{equation}
\xi(\mathbf{v}, \mathbf{t}) = \arccos\left(\frac{\langle \mathbf{v}, \mathbf{t} \rangle_\mathcal{L}}{\|\mathbf{v}\|_\mathcal{L} \|\mathbf{t}\|_\mathcal{L}}\right)
\end{equation}
The entailment constraint loss penalizes text embeddings that fall outside the cone
\begin{equation}
\mathcal{L}_{order} = \max(0, \xi(\mathbf{v}, \mathbf{t}) - \psi(\mathbf{v}))
\end{equation}
This constraint ensures that text representations remain within the semantic cone defined by their corresponding visual content, naturally modeling the containment relationship.

\begin{table*}[t]
\centering
\caption{\textbf{Comparison of AsyMoE with ablated variants on multi-modal benchmarks.} ``Data'' refers to visual instruction data.  Bold numbers indicate the best performance.}
\setlength\tabcolsep{2pt}
\resizebox{1\textwidth}{!}{
\begin{tabular}{ccccccccccccccccc}
\toprule
\multicolumn{1}{c|}{} & \multicolumn{1}{c|}{} & \multicolumn{1}{c|}{} &
MMBench & GQA & VQAv2 & MMVet & \multicolumn{1}{c|}{SEED-I} &
POPE & \multicolumn{1}{c|}{HalluB} &
TextVQA & DocVQA & \multicolumn{1}{c|}{ChartQA} &
AI2D & MMMU$_\text{v}$ & \multicolumn{1}{c|}{Mathvista} & \\
\multicolumn{1}{c|}{\multirow{-2}{*}{}} &
\multicolumn{1}{c|}{\multirow{-2}{*}{Data}} &
\multicolumn{1}{c|}{\multirow{-2}{*}{Backbone}} &
\multicolumn{5}{c|}{General Multi-modal QA} &
\multicolumn{2}{c|}{Hallucination} &
\multicolumn{3}{c|}{OCR-based QA} &
\multicolumn{3}{c|}{Knowledge-based QA} &
\multirow{-2}{*}{AVG} \\
\midrule
\multicolumn{1}{c|}{Dense} 
  & \multicolumn{1}{c|}{2M} 
  & \multicolumn{1}{c|}{Phi-3-mini} 
  & 74.4 & 62.7 & 81.6 & 42.9
  & \multicolumn{1}{c|}{70.2} & 86.2 & \multicolumn{1}{c|}{28.3} 
  & 65.1 & 41.9 & \multicolumn{1}{c|}{53.2} 
  & 64.3 & 41.3 & \multicolumn{1}{c|}{30.8} 
  & 57.1 \\
\multicolumn{1}{c|}{\cellcolor[HTML]{FFFFFF}Vanilla MoE} 
  & \multicolumn{1}{c|}{\cellcolor[HTML]{FFFFFF}2M} 
  & \multicolumn{1}{c|}{\cellcolor[HTML]{FFFFFF}Phi-3-mini} 
  & \cellcolor[HTML]{FFFFFF}75.1 & \cellcolor[HTML]{FFFFFF}62.8 & \cellcolor[HTML]{FFFFFF}81.5 & \cellcolor[HTML]{FFFFFF}42.8 
  & \multicolumn{1}{c|}{\cellcolor[HTML]{FFFFFF}70.6} & \cellcolor[HTML]{FFFFFF}86.3 & \multicolumn{1}{c|}{\cellcolor[HTML]{FFFFFF}28.8} 
  & \cellcolor[HTML]{FFFFFF}65.1 & \cellcolor[HTML]{FFFFFF}42.2 & \multicolumn{1}{c|}{\cellcolor[HTML]{FFFFFF}53.1} 
  & \cellcolor[HTML]{FFFFFF}65.2 & \cellcolor[HTML]{FFFFFF}40.6 & \multicolumn{1}{c|}{\cellcolor[HTML]{FFFFFF}29.9} 
  & \cellcolor[HTML]{FFFFFF}57.2 \\
\multicolumn{1}{c|}{\cellcolor[HTML]{FFFFFF}Modality MoE} 
  & \multicolumn{1}{c|}{\cellcolor[HTML]{FFFFFF}2M} 
  & \multicolumn{1}{c|}{\cellcolor[HTML]{FFFFFF}Phi-3-mini} 
  & \cellcolor[HTML]{FFFFFF}75.1 & \cellcolor[HTML]{FFFFFF}62.9 & \cellcolor[HTML]{FFFFFF}81.7 & \cellcolor[HTML]{FFFFFF}41.0 
  & \multicolumn{1}{c|}{\cellcolor[HTML]{FFFFFF}71.0} & \cellcolor[HTML]{FFFFFF}86.2 & \multicolumn{1}{c|}{\cellcolor[HTML]{FFFFFF}29.3} 
  & \cellcolor[HTML]{FFFFFF}66.0 & \cellcolor[HTML]{FFFFFF}42.7 & \multicolumn{1}{c|}{\cellcolor[HTML]{FFFFFF}53.1} 
  & \cellcolor[HTML]{FFFFFF}65.5 & \cellcolor[HTML]{FFFFFF}40.4 & \multicolumn{1}{c|}{\cellcolor[HTML]{FFFFFF}31.0} 
  & \cellcolor[HTML]{FFFFFF}57.4 \\
\multicolumn{1}{c|}{ContextMoE}            
  & \multicolumn{1}{c|}{2M}   
  & \multicolumn{1}{c|}{Phi-3-mini} 
  & 74.8 & 62.5 & 80.9 & 42.0    
  & \multicolumn{1}{c|}{70.4} & 85.8 & \multicolumn{1}{c|}{29.7} 
  & 64.6 & 42.1 & \multicolumn{1}{c|}{52.5} 
  & 65.0 & 40.6 & \multicolumn{1}{c|}{31.2} 
  & 57.1 \\  
\multicolumn{1}{c|}{MoIIE}            
  & \multicolumn{1}{c|}{2M}   
  & \multicolumn{1}{c|}{Phi-3-mini} 
  & 75.3 & 63.2 & 81.8 & 42.8    
  & \multicolumn{1}{c|}{71.2} & 86.6 & \multicolumn{1}{c|}{30.5} 
  & 65.4 & 42.9 & \multicolumn{1}{c|}{53.3} 
  & 65.9 & 41.3 & \multicolumn{1}{c|}{31.9} 
  & 57.9 \\  
\multicolumn{1}{c|}{\cellcolor[HTML]{E1EAFF}
\textbf{AsyMoE}}           
  & \multicolumn{1}{c|}{\cellcolor[HTML]{E1EAFF}
2M}   
  & \multicolumn{1}{c|}{\cellcolor[HTML]{E1EAFF}
Phi-3-mini} 
  &\cellcolor[HTML]{E1EAFF} \textbf{75.9} &\cellcolor[HTML]{E1EAFF} \textbf{64.7} &\cellcolor[HTML]{E1EAFF} \textbf{82.6} &\cellcolor[HTML]{E1EAFF} \textbf{43.4}    
  & \multicolumn{1}{c|}{\cellcolor[HTML]{E1EAFF}
\textbf{72.4}} &\cellcolor[HTML]{E1EAFF} \textbf{86.6} & \multicolumn{1}{c|}{\cellcolor[HTML]{E1EAFF}
\textbf{31.1}} 
  &\cellcolor[HTML]{E1EAFF} \textbf{66.4} &\cellcolor[HTML]{E1EAFF} \textbf{43.5} & \multicolumn{1}{c|}{\cellcolor[HTML]{E1EAFF}
\textbf{54.7}} 
  &\cellcolor[HTML]{E1EAFF} \textbf{66.8} &\cellcolor[HTML]{E1EAFF} \textbf{42.5} & \multicolumn{1}{c|}{\cellcolor[HTML]{E1EAFF}
\textbf{32.7}} 
  &\cellcolor[HTML]{E1EAFF} \textbf{58.3} \\ 
\midrule
\multicolumn{17}{c}{\cellcolor[HTML]{DEE0E3}
\textbf{\textit{With More Visual Instruction Data}}}
\\ 
\midrule
\multicolumn{1}{c|}{Dense} 
  & \multicolumn{1}{c|}{2.7M} 
  & \multicolumn{1}{c|}{Phi-3-mini} 
  & 75.0 & 63.4 & 81.6 & 41.1 
  & \multicolumn{1}{c|}{70.1} & 87.0 & \multicolumn{1}{c|}{31.7} 
  & 64.6 & 48.2 & \multicolumn{1}{c|}{57.5} 
  & 73.7 & 40.1 & \multicolumn{1}{c|}{31.0} 
  & 58.8 \\
\multicolumn{1}{c|}{Vanilla MoE} 
  & \multicolumn{1}{c|}{2.7M} 
  & \multicolumn{1}{c|}{Phi-3-mini} 
  & 75.2 & 63.3 & 81.4 & 42.2 
  & \multicolumn{1}{c|}{70.6} & 87.2 & \multicolumn{1}{c|}{30.8} 
  & 65.2 & 47.8 & \multicolumn{1}{c|}{57.2} 
  & 74.7 & 42.2 & \multicolumn{1}{c|}{31.2} 
  & 59.1 \\
\multicolumn{1}{c|}{Modality MoE} 
  & \multicolumn{1}{c|}{2.7M} 
  & \multicolumn{1}{c|}{Phi-3-mini} 
  & 73.9 & 63.5 & 81.7 & 40.1 
  & \multicolumn{1}{c|}{70.4} & 87.1 & \multicolumn{1}{c|}{\textbf{32.9}} 
  & 65.7 & 48.9 & \multicolumn{1}{c|}{58.8} 
  & 74.1 & 40.4 & \multicolumn{1}{c|}{31.3} 
  & 59.1 \\
\multicolumn{1}{c|}{ContextMoE}            
  & \multicolumn{1}{c|}{2.7M} 
  & \multicolumn{1}{c|}{Phi-3-mini} 
  & 75.4 & 63.3 & 81.2 & 41.9     
  & \multicolumn{1}{c|}{70.9} & 86.4 & \multicolumn{1}{c|}{29.8} 
  & 65.7 & 47.6 & \multicolumn{1}{c|}{57.3} 
  & 74.8 & 41.5 & \multicolumn{1}{c|}{31.2} 
  & 59.1 \\  
\multicolumn{1}{c|}{MoIIE}            
  & \multicolumn{1}{c|}{2.7M} 
  & \multicolumn{1}{c|}{Phi-3-mini} 
  & 75.9 & 64.1 & 82.0 & 42.7     
  & \multicolumn{1}{c|}{71.8} & 87.2 & \multicolumn{1}{c|}{30.7} 
  & 66.9 & 48.5 & \multicolumn{1}{c|}{58.8} 
  & 75.9 & 42.4 & \multicolumn{1}{c|}{32.5} 
  & 60.0 \\ 
\multicolumn{1}{c|}{\cellcolor[HTML]{E1EAFF}\textbf{AsyMoE}}            
  & \multicolumn{1}{c|}{\cellcolor[HTML]{E1EAFF}2.7M} 
  & \multicolumn{1}{c|}{\cellcolor[HTML]{E1EAFF}Phi-3-mini} 
  &\cellcolor[HTML]{E1EAFF}
 \textbf{76.8} &\cellcolor[HTML]{E1EAFF}
 \textbf{65.3} &\cellcolor[HTML]{E1EAFF}
 \textbf{83.2} &\cellcolor[HTML]{E1EAFF}
 \textbf{43.9}     
  & \multicolumn{1}{c|}{\cellcolor[HTML]{E1EAFF}\textbf{72.6}} &\cellcolor[HTML]{E1EAFF}
 \textbf{88.1} & \multicolumn{1}{c|}{\cellcolor[HTML]{E1EAFF}31.8} 
  &\cellcolor[HTML]{E1EAFF}
 \textbf{68.1} &\cellcolor[HTML]{E1EAFF}
 \textbf{49.7} & \multicolumn{1}{c|}{\cellcolor[HTML]{E1EAFF}\textbf{60.1}} 
  &\cellcolor[HTML]{E1EAFF}
 \textbf{77.3} &\cellcolor[HTML]{E1EAFF}
 \textbf{43.6} & \multicolumn{1}{c|}{\cellcolor[HTML]{E1EAFF}\textbf{33.7}} 
  &\cellcolor[HTML]{E1EAFF}
 \textbf{61.2} \\ 
\midrule
\multicolumn{17}{c}{\cellcolor[HTML]{DEE0E3}\textbf{\textit{With Larger LLM Backbone}}} \\
\midrule
\multicolumn{1}{c|}{Dense} 
  & \multicolumn{1}{c|}{2M} 
  & \multicolumn{1}{c|}{LLaMA3-8B} 
  & 74.8 & 64.8 & 82.2 & 47.9
  & \multicolumn{1}{c|}{72.1} & 86.6 & \multicolumn{1}{c|}{30.1} 
  & 67.2 & 43.4 & \multicolumn{1}{c|}{54.0} 
  & 76.6 & 41.1 & \multicolumn{1}{c|}{30.9} 
  & 59.4 \\
\multicolumn{1}{c|}{Vanilla MoE} 
  & \multicolumn{1}{c|}{2M} 
  & \multicolumn{1}{c|}{LLaMA3-8B} 
  & 75.5 & 64.5 & 82.1 & 46.0 
  & \multicolumn{1}{c|}{71.6} & 86.3 & \multicolumn{1}{c|}{33.0} 
  & 67.3 & 43.8 & \multicolumn{1}{c|}{54.7} 
  & 76.2 & 40.7 & \multicolumn{1}{c|}{30.6} 
  & 59.4 \\
\multicolumn{1}{c|}{Modality MoE} 
  & \multicolumn{1}{c|}{2M} 
  & \multicolumn{1}{c|}{LLaMA3-8B} 
  & 75.6 & 64.3 & 82.3 & 46.3 
  & \multicolumn{1}{c|}{72.3} & 86.7 & \multicolumn{1}{c|}{35.0} 
  & 67.6 & 44.9 & \multicolumn{1}{c|}{55.4} 
  & 76.7 & 41.6 & \multicolumn{1}{c|}{31.2} 
  & 60.0 \\
\multicolumn{1}{c|}{ContextMoE}            
  & \multicolumn{1}{c|}{2M}   
  & \multicolumn{1}{c|}{LLaMA3-8B} 
  & 75.1 & 64.3 & 81.7 & 46.8    
  & \multicolumn{1}{c|}{72.4} & 86.5 & \multicolumn{1}{c|}{35.9} 
  & 67.3 & 44.1 & \multicolumn{1}{c|}{55.4} 
  & 76.2 & 42.1 & \multicolumn{1}{c|}{31.6} 
  & 60.0 \\  
\multicolumn{1}{c|}{MoIIE}            
  & \multicolumn{1}{c|}{2M}   
  & \multicolumn{1}{c|}{LLaMA3-8B} 
  & 75.7 & 64.9 & 82.3 & 47.5    
  & \multicolumn{1}{c|}{73.0} & 87.0 & \multicolumn{1}{c|}{36.5} 
  & 67.9 & 44.7 & \multicolumn{1}{c|}{56.0} 
  & 76.9 & 42.8 & \multicolumn{1}{c|}{32.2} 
  & 60.6 \\
\multicolumn{1}{c|}{\cellcolor[HTML]{E1EAFF}\textbf{AsyMoE}}            
  & \multicolumn{1}{c|}{\cellcolor[HTML]{E1EAFF}2M}   
  & \multicolumn{1}{c|}{\cellcolor[HTML]{E1EAFF}LLaMA3-8B} 
  &\cellcolor[HTML]{E1EAFF} \textbf{76.4} &\cellcolor[HTML]{E1EAFF} \textbf{66.1} &\cellcolor[HTML]{E1EAFF} \textbf{83.9} &\cellcolor[HTML]{E1EAFF} \textbf{48.7}    
  & \multicolumn{1}{c|}{\cellcolor[HTML]{E1EAFF}\textbf{74.3}} &\cellcolor[HTML]{E1EAFF} \textbf{88.4} & \multicolumn{1}{c|}{\cellcolor[HTML]{E1EAFF}\textbf{38.1}} 
  &\cellcolor[HTML]{E1EAFF} \textbf{69.5} &\cellcolor[HTML]{E1EAFF} \textbf{46.2} & \multicolumn{1}{c|}{\cellcolor[HTML]{E1EAFF}\textbf{57.6}} 
  &\cellcolor[HTML]{E1EAFF} \textbf{78.5} &\cellcolor[HTML]{E1EAFF} \textbf{44.3} & \multicolumn{1}{c|}{\cellcolor[HTML]{E1EAFF}\textbf{33.7}} 
  &\cellcolor[HTML]{E1EAFF} \textbf{62.1} \\  
\bottomrule
\end{tabular}}
\label{tab:2}
\end{table*}

\begin{figure*}[t]
   \centering
   \includegraphics[width=\textwidth]{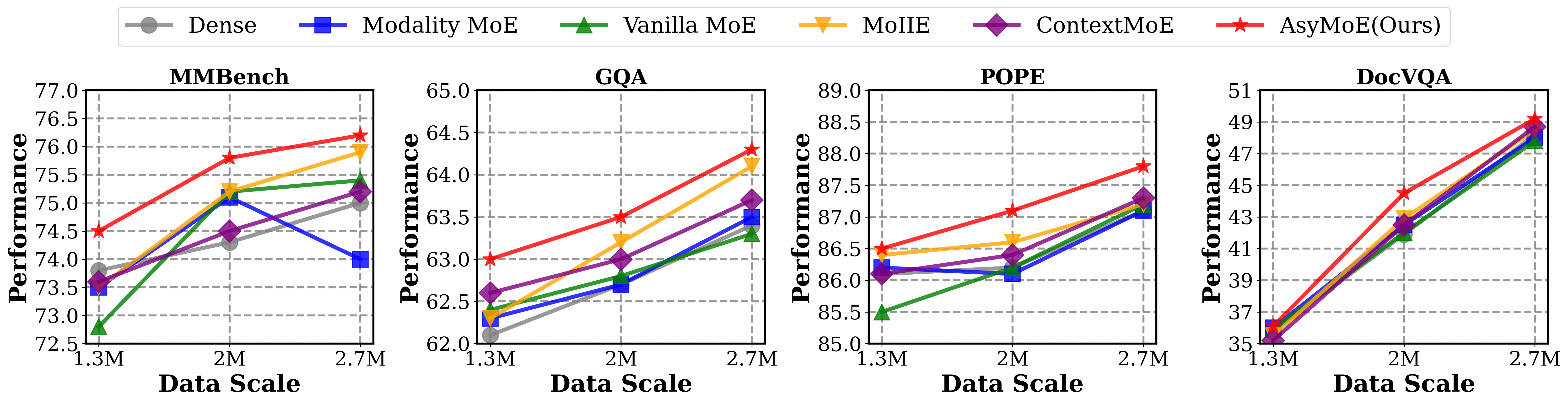}
   \caption{\textbf{Performance scaling across visual instruction tuning data scales.} AsyMoE demonstrates superior scaling efficiency compared to dense and alternative MoE architectures.}
   \label{fig:4}
\end{figure*}

\begin{figure*}[t]
   \centering
   \includegraphics[width=\textwidth]{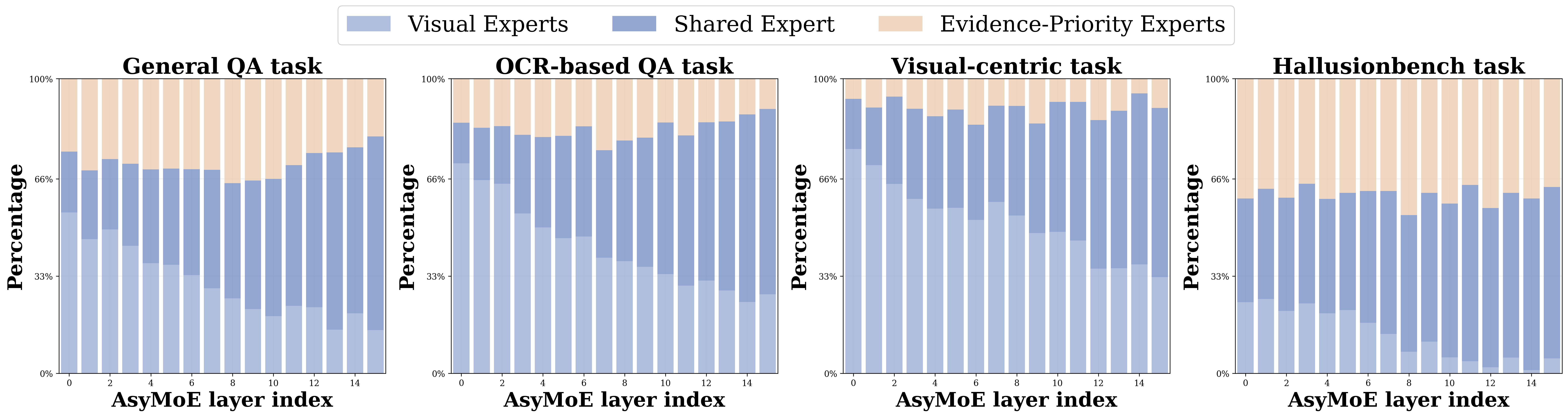}
   \caption{\textbf{Expert Activation Distribution Across Tasks.}  We visualize the dynamic distribution of activated experts across various multimodal tasks, reflecting how modality-specific and context-aware experts contribute to different reasoning challenges.}
   \label{fig:5}
\end{figure*}


\subsection{Routing Strategy with Modal Awareness}
The routing mechanism in AsyMoE considers both modality characteristics and evidence relevance. For visual tokens, the routing probability is computed as
\begin{equation}
g_V(\mathbf{h}_v) = \text{Softmax}(\mathbf{W}_V \cdot \mathbf{h}_v)
\end{equation}
where $\mathbf{W}_V$ is the routing weight matrix for visual tokens. Visual tokens are routed to visual intra-modality experts $\mathcal{E}_V$ and shared inter-modality experts $\mathcal{E}_S$.

For language tokens, the routing strategy incorporates evidence awareness
\begin{equation}
g_L(\mathbf{h}_l) = \text{Softmax}(\mathbf{W}_L \cdot \mathbf{h}_l + s_{evd} \cdot \mathbf{m}_{evd})
\end{equation}
where $\mathbf{W}_L$ is the routing weight matrix for language tokens. $\mathbf{m}_{evd} \in \mathbb{R}^{|\mathcal{E}|}$ is a binary mask vector with value 1 for evidence-priority expert positions and 0 otherwise. $s_{evd}$ is the evidence relevance score computed as
\begin{equation}
s_{evd} = \sigma(\mathbf{w}_{evd}^T \cdot \text{Attn}(\mathbf{h}_l, \mathbf{c}))
\end{equation}
Here $\sigma$ is the sigmoid function. $\mathbf{w}_{evd} \in \mathbb{R}^d$ is a learnable parameter. $\text{Attn}(\mathbf{h}_l, \mathbf{c})$ computes the attention-weighted context representation. When the input context is highly relevant to the current token, $s_{evd}$ increases, boosting the routing probability toward evidence-priority experts. This mechanism ensures that tokens requiring contextual grounding are directed to appropriate experts.

The overall training objective combines the language modeling loss with auxiliary losses
\begin{equation}
\mathcal{L} = \mathcal{L}_{lm} + \lambda_1 \mathcal{L}_{order} + \lambda_2 \mathcal{L}_{balance}
\end{equation}
where $\mathcal{L}_{lm}$ is the standard cross-entropy loss for next token prediction. $\mathcal{L}_{order}$ is the entailment cone constraint. $\mathcal{L}_{balance}$ is the load balancing loss for MoE training. $\lambda_1$ and $\lambda_2$ are weighting coefficients.

\section{EXPERIMENTS}
\subsection{Experimental Setup}
We evaluate AsyMoE on 18 multimodal benchmarks covering general understanding with \textbf{MMBench-EN} \cite{liu_2024_mmbench}, \textbf{MM-Vet} \cite{pmlr-v235-yu24o}, \textbf{GQA} \cite{Hudson_2019_CVPR}, \textbf{VQAv2}, and \textbf{SEED-Image} \cite{Li_2024_CVPR}, knowledge-based QA using \textbf{MMMU} \cite{yue2024mmmumassivemultidisciplinemultimodal}, AI2D \cite{kembhavi-2016-diagram}, SciQA-IMG \cite{NEURIPS2022_11332b6b}, and MathVista \cite{ICLR2024_663bce02}, OCR tasks through TextVQA \cite{Singh_2019_CVPR}, ChartQA \cite{masry-etal-2022-chartqa}, and DocVQA \cite{Mathew_2021_WACV}, hallucination robustness via POPE \cite{li-etal-2023-evaluating} and HallusionBench \cite{Guan_2024_CVPR}, and context-dependent reasoning with SQuAD, NQ, HotpotQA, NQ-Swap, and ConfiQA. We compare against dense models without MoE, MoE variants including Vanilla MoE with shared experts and Modality-specific MoE with separate experts, state-of-the-art models like \textbf{SPHINX-X} \cite{liu2024sphinxxscalingdataparameters}, \textbf{MM1} \cite{mckinzie_2024_mm1}, and \textbf{CuMO}, and context utilization methods like \textbf{ContextMoE} \cite{bai-etal-2025-understanding}.

\subsection{Implementations}
We implement AsyMoE using phi-3-mini and LLaMA3-8B as LLM backbones with a pre-trained SigLIP visual encoder and two-layer MLP connection module. The architecture employs a 4-expert configuration comprising intra-modality experts for vision and language, inter-modality experts for cross-modal interactions, and evidence-priority language experts for suppressing parametric biases. Training follows a two-stage strategy where the first stage optimizes the connection module on Bunny-pretrain-LAION-2M dataset, and the second stage integrates the complete AsyMoE architecture with learning rates of $2 \times 10^{-6}$ for visual encoder and $2 \times 10^{-5}$ for other components on visual instruction datasets. All experiments use AdamW optimizer with cosine scheduling and are conducted on 8 NVIDIA H20 GPUs using DeepSpeed ZeRO-3 for efficient training.

\subsection{Main Results}

\textbf{Comparison with State-of-the-art Models.} Table~\ref{tab:1} presents the performance comparison between AsyMoE and existing LVLMs across diverse multimodal benchmarks. We organize models by parameter scale and architectural design for fair comparison. AsyMoE-Phi3 achieves 68.2\% on TextVQA and 75.9\% on MMBench with only 4.1B activated parameters. These results outperform MoIIE-Phi3 (65.4\% and 75.4\%) and ContextMoE-Phi2 (66.2\% and 74.9\%) with fewer computational resources. When scaling to a larger backbone, AsyMoE-LLaMA3 reaches 70.9\% on TextVQA and 76.8\% on MMBench with 10.8B activated parameters. This surpasses CuMo built on Mixtral-8$\times$7B, requiring 13.5B activated parameters. The consistent improvements across different backbones demonstrate the generality of our approach.

\noindent\textbf{Comparison with MoE Variants.} Table~\ref{tab:2} provides controlled comparisons against MoE architectural variants under identical training configurations. AsyMoE consistently outperforms Vanilla MoE, Modality MoE, ContextMoE, and MoIIE across all benchmark categories. The improvements are particularly notable on hallucination benchmarks. AsyMoE achieves 31.1\% on HallusionBench compared to 30.5\% for MoIIE and 28.8\% for Vanilla MoE under the 2M data setting. This 2.3\% absolute improvement over Vanilla MoE validates the effectiveness of evidence-priority experts in maintaining contextual grounding. On knowledge-based QA benchmarks requiring integration of visual evidence with reasoning, AsyMoE achieves 42.5\% on MMMU and 32.7\% on MathVista, outperforming MoIIE by 1.2\% and 0.8\% respectively. The average improvement across 13 benchmarks is 1.5\% over MoIIE and 1.9\% over Vanilla MoE.

\noindent\textbf{Training Dynamics.} Figure~\ref{fig:3} illustrates the training behavior of AsyMoE compared to baseline architectures. The left panel shows that AsyMoE achieves smoother loss convergence throughout training. The right panel demonstrates that AsyMoE reaches higher validation accuracy earlier in training and maintains this advantage. At 1200 training steps, AsyMoE achieves approximately 80\% accuracy on MMBench validation, compared to 75\% for Modality MoE and 70\% for Vanilla MoE. This faster convergence suggests that the asymmetry-aware design provides more effective learning signals.


\noindent\textbf{Scaling Behavior.} Figure~\ref{fig:4} studies performance under different visual instruction tuning data scales. AsyMoE improves consistently from 1.3M to 2.7M samples, while Dense and Modality MoE show weaker gains beyond 2M samples. The advantage is most evident on DocVQA, where AsyMoE improves from 39\% to 49\%, compared with a 6\% gain for Vanilla MoE. This confirms that modeling modal asymmetry enables more effective use of additional training data.

\subsection{Expert Behavior Validation}
\noindent\textbf{Task-Adaptive Expert Activation.} Figure~\ref{fig:5} shows that expert activation is strongly task-dependent. On General QA and hallucination benchmarks, evidence-priority experts dominate in deeper layers, indicating that AsyMoE preserves contextual grounding when faithful evidence use is critical. On OCR-based QA, visual experts remain more active throughout the network, reflecting the importance of modality-specific visual processing. For visual-centric tasks, activations are more balanced between visual experts and shared inter-modality experts, suggesting effective cross-modal integration. Overall, these patterns confirm that AsyMoE adaptively routes tokens according to task demands rather than relying on a fixed expert preference.

\begin{figure}[t]
   \centering
   \includegraphics[width=\columnwidth]{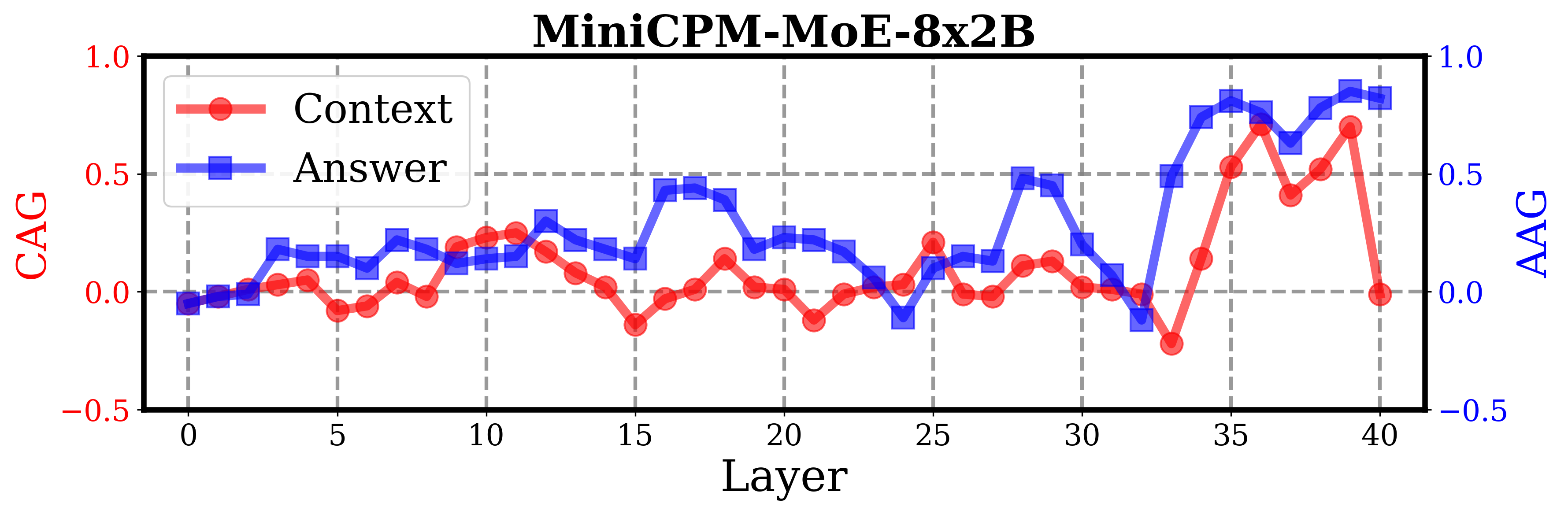}
   \caption{\textbf{Layer-wise Attention Gain.} 
   }
   \label{fig:6}
\end{figure}

\noindent\textbf{Layer-wise Attention Analysis.} Figure~\ref{fig:6} presents the Context Attention Gain (CAG) and Answer Attention Gain (AAG) metrics across network layers. AsyMoE maintains stable Context Attention Gain across layers, while baseline models show a clear decline in deeper layers, indicating context dilution. Meanwhile, Answer Attention Gain in AsyMoE increases progressively with depth, this pattern indicates that AsyMoE successfully maintains contextual grounding in early layers and progressively focuses on answer-relevant information in later layers.

\begin{table}[!t] 
\footnotesize 
\centering 
\caption{Memory-evidence balance parameter sensitivity analysis.} 
\setlength{\tabcolsep}{2.10mm} 
\setlength\heavyrulewidth{0.25ex} 
\renewcommand{\arraystretch}{1.0} 
\begin{tabular}{@{}ccccccc@{}} 
\toprule 
$\alpha$ & TextVQA        & MMBench        & GQA            & MMMU$_\text{v}$ & POPE           & MMVet          \\ 
\midrule 
0.1   & 67.32          & 75.41          & 64.85          & 41.89          & 87.28          & 44.62          \\ 
0.3   & \textbf{68.24} & \textbf{75.94} & \textbf{65.18} & \textbf{42.31} & \textbf{87.64} & \textbf{45.83} \\ 
0.5   & 67.85          & 75.67          & 64.93          & 42.07          & 87.45          & 45.29          \\ 
0.7   & 66.94          & 75.12          & 64.52          & 41.74          & 87.11          & 44.56          \\ 
\bottomrule 
\end{tabular} 
\label{tab:3} 
\end{table}

\begin{table}[!t]
    \centering
        \caption{Ablation study results with Phi-3-mini backbone.}
    \setlength\heavyrulewidth{0.25ex} 
    \resizebox{\columnwidth}{!}{
    \begin{tabular}{lcccc}
    \toprule
         Method & TextVQA$\uparrow$ & MMBench$\uparrow$ & GQA$\uparrow$ & AVG$\uparrow$ \\
         \midrule
         \cellcolor{blue!15}\textbf{AsyMoE (Full)} & \cellcolor{blue!15}\textbf{68.1} & \cellcolor{blue!15}\textbf{76.8} & \cellcolor{blue!15} \textbf{65.3} & \cellcolor{blue!15} \textbf{61.2}\\
         \textit{- w/o Hyperbolic Space} & 66.8 & 75.1 & 64.2 & 58.7\\
         \textit{- w/o Evidence-Priority Experts} & 67.1 & 74.9 & 64.5 & 58.8\\
         \textit{- w/o Cross-Modal Experts} & 67.5 & 75.2 & 64.8 & 59.2\\
         \textit{- w/o Intra-modality Separation} & 67.0 & 74.8 & 64.3 & 58.7\\
         
         \bottomrule
    \end{tabular}
    }
    \label{tab:4}
\end{table}

\subsection{Ablation Studies}
Table~\ref{tab:4} reports the contribution of each component. Removing evidence-priority experts causes the largest drop, reducing average accuracy from 61.2\% to 58.8\%, indicating their key role in preserving evidence grounding. Excluding hyperbolic space also degrades performance, especially on TextVQA and MMBench, confirming its benefit for cross-modal relation modeling. Removing cross-modal experts or intra-modality separation further reduces accuracy, showing that both designs are necessary for effective multimodal interaction. Table~\ref{tab:3} analyzes the memory-evidence balance parameter $\alpha$. The best performance appears at $\alpha=0.3$, and results remain stable from $\alpha=0.1$ to $\alpha=0.5$. A larger value of $\alpha=0.7$ leads to a clearer decline, suggesting that evidence priority should not be overly weakened by memory suppression.

\section{Conclusion}
We presented AsyMoE, an MoE architecture designed to model the asymmetry between visual and linguistic processing. To address memory-priority shift and insufficient cross-modal structure modeling, AsyMoE introduces evidence-priority language experts and hyperbolic inter-modality experts. Experiments on 18 benchmarks show consistent improvements, especially on hallucination-sensitive tasks. Future work will extend this design to other modalities such as audio and video.

\begin{acks}
    This research was supported by Oil \& Gas Major Project (No. 2025ZD1404600), CNPC Innovation Found (No. 2022DQ02-0609), and Frontier Interdisciplinary Exploration Research Program of China University of Petroleum, Beijing (Grant No. 2462024XKQY003).
\end{acks}

\newpage
\bibliographystyle{ACM-Reference-Format}
\bibliography{main}

\end{document}